\documentclass[akbc,twoside,11pt,lettersize]{article}
\usepackage{akbc}
\akbcheading
\ShortHeadings{Syntactic Question Abstraction and Retrieval}{Hwang, Yim, Park, \& Seo}
\finalcopy 

\usepackage{times}
\usepackage{url}
\usepackage{longtable}
\usepackage{multirow}
\usepackage{lineno}

\usepackage{latexsym}
\usepackage{threeparttable}
\usepackage{amsmath,amssymb,bm,color,booktabs}
\usepackage{todonotes}
\usepackage{tabularx}

\newcommand{\Ours}{\textsc{Syntactic Question Abstraction \& Retrieval}}
\newcommand{\ours}{\textsc{SQAR}}

\newcommand{\sqlova}{\textsc{SQLova}}
\newcommand{\sqlovaglove}{\textsc{SQLova-GloVe}}

\newcommand{\ctof}{\textsc{Coarse2Fine}}

\newcommand{\tusset}{Test-Uniform-81P-648}
\newcommand{\tfset}{Test-Full-15878}

\newcommand{\usset}{Train-Uniform-85P-850}  
\newcommand{\ussetd}{Dev-Uniform-80P-320}  
\newcommand{\umset}{Train-Uniform-85P-2550} 
\newcommand{\umsetd}{Dev-Uniform-80P-320}

\newcommand{\ufset}{Train-Full-56355} 

\newcommand{\rsset}{Train-Rand-881}
\newcommand{\rssetd}{Dev-Rand-132}

\newcommand{\rmset}{Train-Rand-3523} 
\newcommand{\rrmset}{Train-Rand-2677} 

\newcommand{\rmsetd}{Dev-Rand-527}

\newcommand{\hmset}{Train-Hybrid-85P-2670} 
\newcommand{\hhhhmset}{Train-Hybrid-96P-2750} 

\newcommand{\hmsetd}{Dev-Hybrid-446}
\newcommand{\hsset}{Train-Hybrid-85P-897} 
\newcommand{\hssetd}{Dev-Hybrid-223}

\newcommand{\bea}[1]{ \begin{equation}\begin{aligned} #1 \end{aligned} \end{equation} }

\newcommand{\softmax}{\text{softmax}}
\newcommand{\mcW}{\mathcal{W}}

\DeclareMathOperator*{\argmin}{argmin}

\graphicspath{{./figures/}}

\begin{document}

\title{Syntactic Question Abstraction and Retrieval \\for Data-Scarce Semantic Parsing}

\author{\name Wonseok Hwang \email wonseok.hwang@navercorp.com \\
       \name Jinyeong Yim \email jinyeong.yim@navercorp.com \\
       \name Seunghyun Park \email seung.park@navercorp.com \\
       \name Minjoon Seo \email minjoon.seo@navercorp.com \\
       \addr Clova AI, NAVER Corp.
       }


\maketitle

\begin{abstract}
Deep learning approaches to semantic parsing require a large amount of labeled data, but annotating complex logical forms is costly.
  Here, we propose \Ours\ (\ours), a method to build a neural semantic parser that translates a natural language (NL) query to a SQL logical form (LF) with less than 1,000 annotated examples.
  \ours\ first retrieves a logical pattern from the train data by computing the similarity between NL queries and then grounds a lexical information on the retrieved pattern in order to generate the final LF.
  We validate \ours\ by training models using various small subsets of WikiSQL train data achieving up to 4.9\% higher LF accuracy compared to the previous state-of-the-art models on WikiSQL test set. 
  We also show that by using query-similarity to retrieve logical pattern, \ours\ can leverage a paraphrasing dataset achieving up to 5.9\% higher LF accuracy compared to the case where \ours\ is trained by using only WikiSQL data.
  In contrast to a simple pattern classification approach, \ours\ can generate unseen logical patterns upon the addition of new examples without re-training the model.
  We also discuss an ideal way to create cost efficient and robust train datasets when the data distribution can be approximated under a data-hungry setting.
\end{abstract}

\section{Introduction}
\label{Introduction}
Semantic parsing is the task of translating natural language into machine-understandable formal logical forms. 
With the help of recent advance in deep learning technology, neural semantic parsers have achieved state-of-the-art results in many tasks \cite{dong2016Neural_semantic_parsing,jia2016neuralSP_dataAug,iyer2017UserInLoop}.
However, their training requires the preparation of a large amount of labeled data (questions and corresponding logical forms) which is often not scalable due to the requirement of expert knowledge necessary in writing logical forms.

Here, we develop a novel approach \Ours~(\ours) for semantic parsing task under data-hungry setting. The model constrains the logical form search space by retrieving logical patterns from the train set using natural language similarity with assistance of a pre-trained language model. The subsequent grounding module only needs to map the retrieved pattern to the final logical form. 

We evaluate \ours\ on various subsets of WikiSQL train data \cite{zhongSeq2SQL2017} consisting of 850$\sim$2750 samples which occupies 1.5--4.9\% of the full train data. 
\ours\ shows up to 4.9\% higher logical form accuracy compared to the previous best open sourced model \sqlova\ \cite{hwang2019sqlova}.
Also, we show that natural language sentence similarity dataset can be leveraged in \ours\ by pre-training the backbone of \ours\ using Quora pharaphrasing data which results in up to 5.9\% higher logical form accuracy.

In general, the retrieval approach causes the limitation on dealing with unseen logical patterns. In contrast, we show that \ours\ can generate unseen logical patterns by collecting new examples without re-training opening an interesting possibility of generalizable retrieval-based semantic parser.


Our contributions are summarized as follows:
\begin{itemize}
    \item Compared to the previous best open-sourced model~\cite{hwang2019sqlova}, \ours\ achieves the state-of-the-art performance on the WikiSQL test data under data-scarce environment.
    \item We show that \ours\ can leverage natural language query similarity datasets to improve logical form generation accuracy.
    \item We show that retrieval-based parser can handle unseen new logical patterns on the fly without re-training.
    \item For the maximum cost-effectiveness, we find that it is important to carefully design the train data distribution, not merely following the (approximated) data distribution.
\end{itemize}

\section{Related work}
WikiSQL \cite{zhongSeq2SQL2017} is a large semantic parsing dataset consisting of 80,654 natural language utterances and corresponding SQL annotations. Its massive size has invoked the development of many neural semantic parsing models  \cite{xuSQLNet2017,yu2018TypeSQL,dongC2F2018,wang2017pointingOut,wang2018srllike,mcCann2018decaNLP,shi2018IncSQL,yin2018TRANX,xiong2018gan,hwang2019sqlova,he2019xsql}.
Berant and Liang \cite{berant-liang-2014-semantic} built the semantic parser that uses the query similarity between an input question and paraphrased canonical natural language representations generated from candidate logical forms. In our study, candidate logical forms and corresponding canonical forms do not need to be generated as input questions are directly compared to the questions in the training data, circumventing the burden of full logical form generation.
Dong and Lapata \cite{dongC2F2018} developed the two step approach for logical form generation, similar to \ours\ using sketch representation as intermediate logical forms. In \ours, intermediate logical forms are retrieved from train set using question similarity being specialized for data-hungry setting. 
Finegan-Dollak et al. \cite{fineganDollak2018ACLmethodology} developed the model that first finds corresponding logical pattern and fills the slots in the template. While their work resembles \ours, there is a fundamental difference between two approaches. The model from \cite{fineganDollak2018ACLmethodology} \emph{classifies} input query into logical pattern whereas we use query-to-query similarity to \emph{retrieve} logical pattern non-parametrically. By retrieving logical pattern using the similarity in natural language space, paraphrasing datasets can be employed during training which is relatively easy to label compared to semantic parsing datasets. Also, in contrast to classification methods, \ours\ can handle unseen logical patterns by including new examples into the train set \emph{without} re-training the model during inference stage (see section. \ref{sec:generalization}). Also our focus is developing competent model with small amount of data which has not been studied in \cite{fineganDollak2018ACLmethodology}.
Hwang et al. \cite{hwang2019sqlova} developed \sqlova\ that achieves state-of-the-arts result in the WikiSQL task. \sqlova\ consits of table-aware BERT encoder and NL2SQL module that generate SQL queries via slot-filling approach.


\section{Model}
\begin{figure}[tb]
\centering
\includegraphics[width=0.6\textwidth]{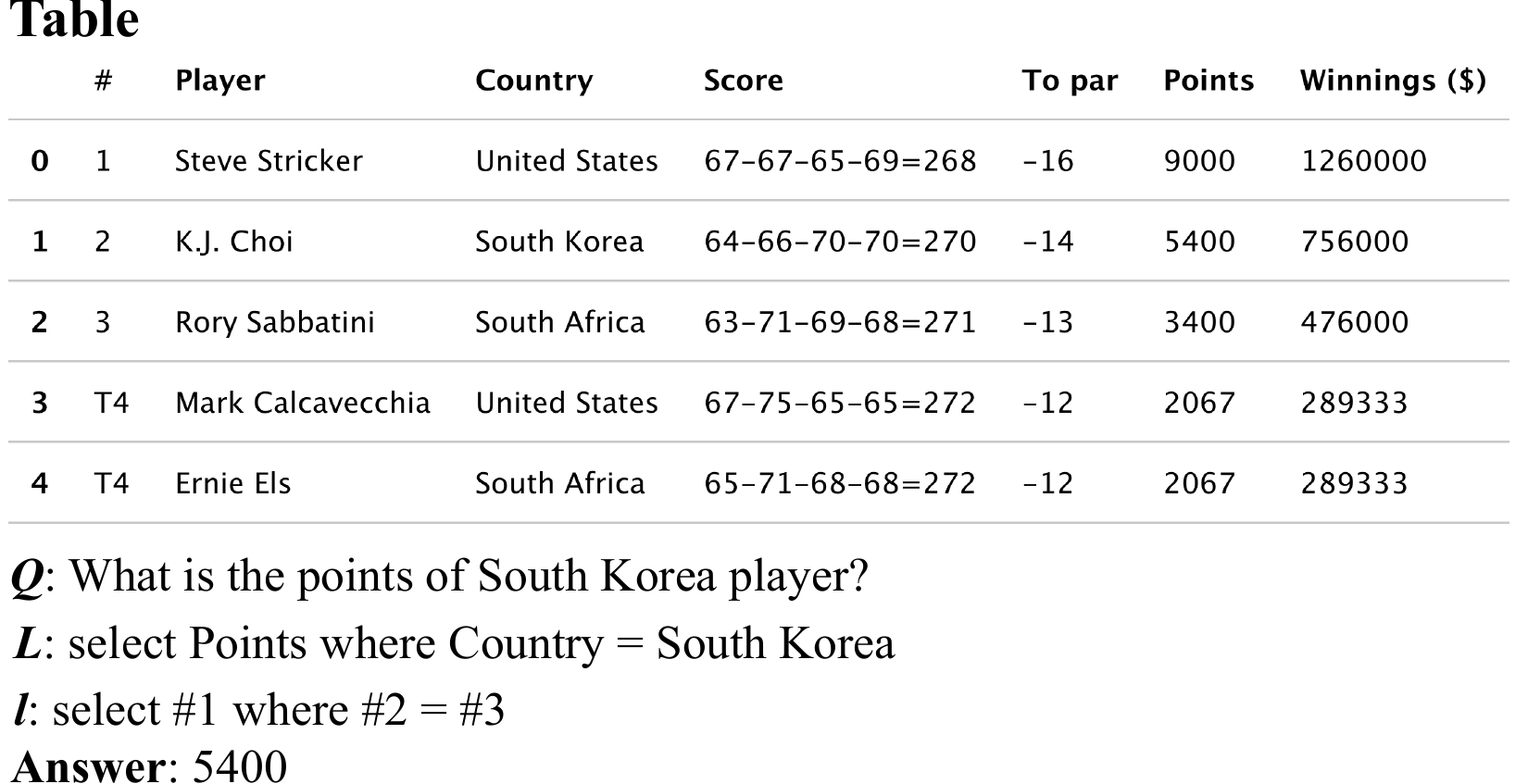}
\caption{Example of WikiSQL semantic parsing task. For given question ($Q$) and table headers, the model generates corresponding SQL query ($L$) and retrieves the answer from the table.
}
\label{fig_intro}
\end{figure}

\setlength{\textfloatsep}{5.0pt}
The model generates the logical form $L$ (SQL query) for a given NL query $Q$ and its corresponding table headers $H$ (Fig. \ref{fig_intro}). First, the logical pattern $l$ is retrieved from the train set by finding the most similar NL query with $Q$. 
For example in Fig. \ref{fig_intro}, $Q$ is ``What is the points of South Korea player?''. To generate logical form $L$, \ours\ retrieves logical pattern $l$ = \texttt{SELECT \#1 WHERE \#2 = \#3} by finding the most similar NL query from the train set, for instance [``Which fruit has yellow color?'', \texttt{SELECT Fruit WHERE Color = Yellow}]. Then \#1, \#2, and \#3 in $l$ are grounded to \texttt{Point}, \texttt{Country}, and \texttt{South Korea} respectively by the grounding module using information from $Q$ and table headers. The process is depicted schematically in Fig. \ref{fig_model}a.
The detail of each step is explained below.


\subsection{ Syntactic Question Abstractor } 
The syntactic question abstractor generates two vector representation $q$ and $g$ of an input NL query $Q$ (Fig. \ref{fig_model}b).
$q$ is trained to represent syntactic information of $Q$ and used in the retriever module (Fig. \ref{fig_model}c).
$g$ is trained to represent lexical information of $Q$ by being used in the grounder (Fig. \ref{fig_model}d).

The logical patterns of the WikiSQL dataset consist of combination of six aggregation operators (\texttt{none}, \texttt{max}, \texttt{min}, \texttt{count}, \texttt{sum}, and \texttt{avg}), and three \texttt{where} operators (\texttt{=}, \texttt{>}, and \texttt{<}). The number of conditions in \texttt{where} clause is ranging from 0 to 4. Each condition is combined by \texttt{and} unit. In total, there are 210 possible SQL patterns (6 \texttt{select} clause patterns  $\times$ 35 \texttt{where} clause patterns, see Fig. \ref{fig_pattern}).
To extract these syntactic information from NL query, both an input NL query $Q$ and the queries in train set $\{Q_{t,i}\}$ are mapped to a vector space (represented by $q$ and $\{q_{t,i}\}$, respectively) via table-aware BERT encoder \cite{devlinBERT2018,hwang2019sqlova} (Fig. \ref{fig_model}b).
The input of the encoder consists of following tokens:
\\\\
\texttt{[CLS]}, $E$, \texttt{[SEP]}, $Q$, \texttt{[SEP]}, $H$, \texttt{[SEP]}
\\\\
where $E$ stands for SQL language element tokens such as \texttt{[SELECT]}, \texttt{[MAX]}, \texttt{[COL]}, $\cdots$) separated by  \texttt{[SEP]} (a special token in BERT), $Q$ represents question tokens, and $H$ denotes the tokens of table headers in which each header is separated by \texttt{[SEP]}. $E$ is included to contextualize and use them during grounding process (section \ref{sec:aux_g}).
Segment ids are used to distinguish $E$ (id = 0) from $Q$ (id = 1) and $H$ (id = 1) as in BERT \cite{devlinBERT2018}.
Next, two vectors $q \equiv v^{\texttt{[CLS]}}_{0:d_q}$ and $g \equiv v^{\texttt{[CLS]}}_{d_q:(d_q+2 d_h)}$ are extracted from the (linearly projected) encoding vector of \texttt{[CLS]} token where $i:j$ notation indicates the elements of vector between $i$th and $j$th indices. In this study, $d_q=256$ and $d_h=100$.

\subsection{ Retriever }
To retrieve logical pattern of $Q$,  
the questions from the train set ($\{Q_{t,i}\}$) are also mapped to the vector space ($\{q_{t,i}\}$) using the syntactic question abstractor. Next, the logical pattern is found by measuring Euclidean $L_2$ distance between $q$ and $\{q_{t,i}\}$.
\bea{ 
    q_{t,i^*} &= \argmin_{q_{t,i}} ||q - q_{t,i}||_{L_2}
}
Since $q_{t,i^*}$ has corresponding $Q_{t,i^*}$ and logical form $L_{t,i^*}$, the logical pattern $l^*$ can be obtained from $L_{t,i^*}$ after delexicalization. The process is depicted in Fig. \ref{fig_model}c.
In \ours, maximum 10 closest $q_{t,i^*}$ are retrieved and the most frequently appearing logical pattern is selected for the subsequent grounding process.
\ours\ is trained using the negative sampling method. First, one positive sample (having the same logical pattern with input query $Q$), and 5 negative samples (having different logical pattern) are randomly sampled from the train set. Then six $L_2$ distances are calculated as above and interpreted as approximate probability by using softmax function after multiplied by -1. The cross entropy function is employed for the training.

\subsection{Grounder} \label{sec:aux_g}
\begin{figure}[t!] 
	\centering
	\includegraphics[width=0.9\textwidth]{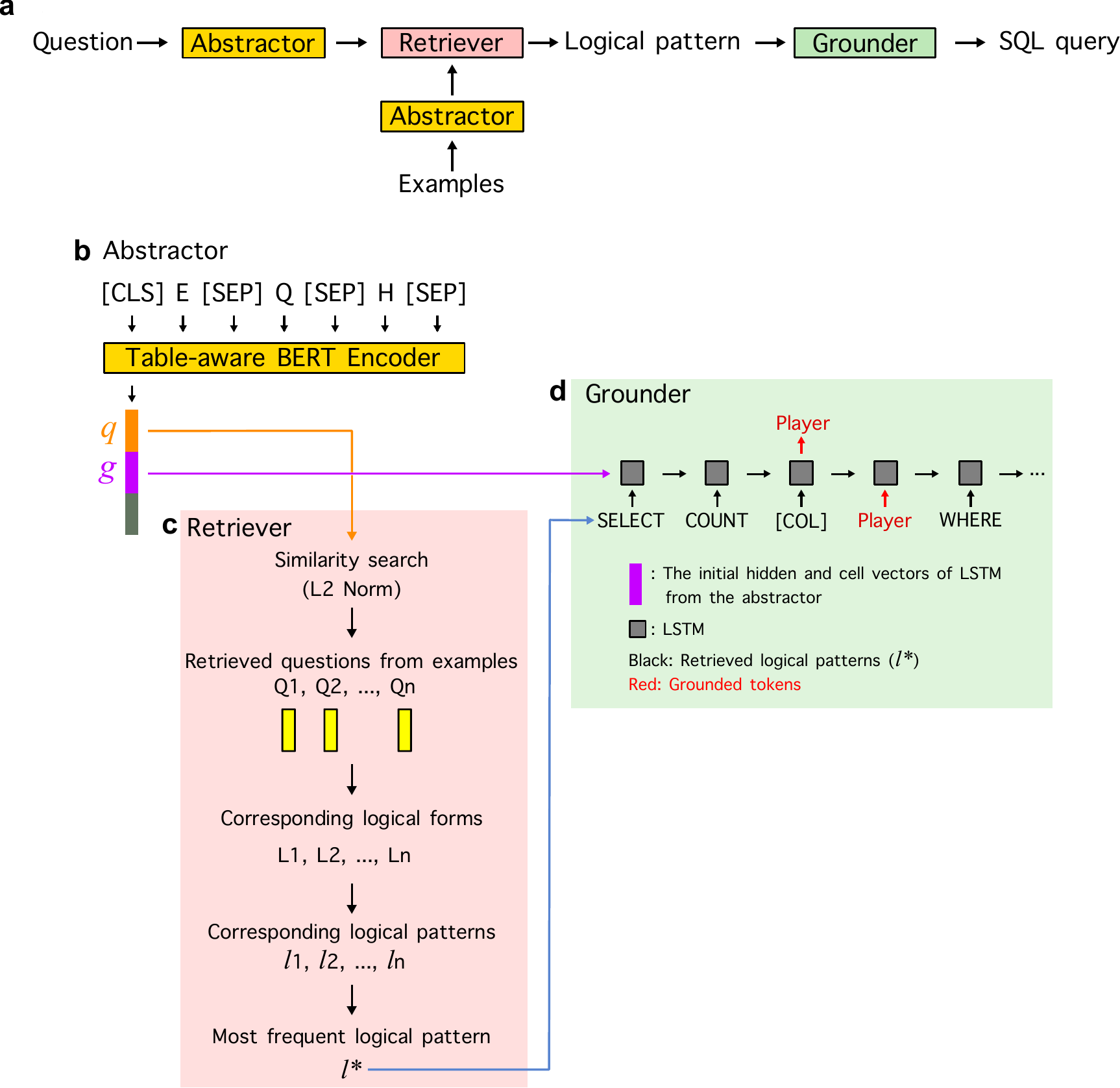}
	\caption{ 
	(a) The schematic representation of \ours. 
	(b) The scheme of the syntactic question abstractor.
	(c) The retriever.
	(d) The grounder. Only lexical tokens (red-colored) are predicted and used as the next input token.
	}
	\label{fig_model}
\end{figure}

To ground retrieved logical pattern $l^*$, following LSTM-based pointer network is used \cite{vinyals2015NIPS_pointerNetwork}. 
\bea{
    D_t &= \text{LSTM}(P_{t-1}, (h_{t-1}, c_{t-1}))
    \\ h_0 &= g_{0:d_h}
    \\ c_0 &= g_{d_h:2d_h}
    \\ s_{t}(i) &=  \mcW ( \mcW H_{i} + \mcW D_t)
    \\ p_{t}(i) &= \softmax ~ s_{t}(i),
}
where $P_{t-1}$ stands for the one-hot vector (pointer to the input token) at time $t-1$, $h_{t-1}$ and $c_{t-1}$ are hidden- and cell-vectors of the LSTM decoder, $\mcW$'s denote (mutually different) affine transformations, and $p_t(i)$ is the probability of observing $i$th input token at time $t$.
Here $d_h$ (=100) is the hidden dimension of the LSTM.
Compared to a conventional pointer network, our grounder has three custom properties: (1) as logical pattern is already found from the retriever, the grounder does not feed the output as the next input when the input token is already present in the logical pattern whereas lexical outputs like column and \texttt{where} values are fed into the next step as an input (Fig. \ref{fig_model}d); (2) to generate conditional values for \texttt{where} clause, the grounder infers only the beginning and the end token positions from the given question to extract the condition values for \texttt{where} clause; (3) the multiple generation of same column on \texttt{where} clause is avoided by constraining the search space.
The syntactic question abstractor, the retriever, and the grounder are together named as \Ours\ (\ours).

\section{Experiments}
To train \ours\ and \sqlova, the pytorch version of pre-trained BERT model\footnote{https://github.com/huggingface/transformers} (\texttt{BERT-Base-Uncased}\footnote{https://github.com/google-research/bert}) is loaded and fine-tuned using ADAM optimizer. The NL query is first tokenized by using Standford CoreNLP~\cite{manningCoreNLP2014}.
Each token is further tokenized (into sub-word level) by WordPiece tokenizer~\cite{devlinBERT2018,wu2016word_piece}. 
FAISS \cite{johnson2017faiss} is employed for the retrieval process.
For the experiments with \usset, \rsset, \hsset, and \rmset, only single logical pattern is retrieved from the retriever due to the scarcity of examples per pattern. Otherwise 10 logical patterns are retrieved. 
All experiments were performed with WikiSQL ver. 1.1 \footnote{https://github.com/salesforce/WikiSQL}.
The accuracy is measured by repeating three independent experiments in each condition with different random seeds unless particularly mentioned.
To further pre-train BERT-backbone of \ours, we use Quora paraphrase detection dataset \cite{quoradata}.
The further details of experiments are summarized in Appendix.

\section{Result and Analysis}
\subsection{ Preparation of data scarce environment }
The WikiSQL dataset consists of 80,654 examples (56,355 in train set, 8,421 in dev set, and 15,878 in test set).
The examples are not uniformly distributed over 210 possible SQL logical patterns in train, dev, and test sets while they have similar logical pattern distributions (see Fig. \ref{fig_pattern}, Table~\ref{tab:pattern_count}).
To mimic original pattern distribution while preparing data scarce environemnts, we prepare \rsset\ by randomly sampling 881 examples from the original WikiSQL train set (1.6\%). The validation set \rssetd\ is prepared by the same way from the WikiSQL dev set.

\subsection{ Accuracy Measurement }
\ours\ retrieves SQL logical pattern for a given question $Q$ by finding most syntactically similar question from the train set and ground the retrieved logical pattern using LSTM-based grounder (Fig. \ref{fig_model}a).
The model performance is tested over the full WikiSQL test set by using two metrics: (1) logical pattern accuracy (P) and (2) logical form accuracy (LF).
P is computed by ignoring difference in lexical information such as predicted columns and conditional values whereas LF is calculated by comparing full logical forms. The execution accuracy of SQL query is not compared as different logical forms can generate identical answer hindering fair comparison.
Table~\ref{tbl_rsset} shows P and LF of several models over the WikiSQL original test set conveying following important messages: (1) \ours\ outperforms \sqlova\ by +4.0\% in LF (3rd and 4th rows); (2) Quora pre-training improves the performance of \ours\ further by 0.9\% (4th and 5th rows); (3) Under data-scarce condition, the use of pre-trained language model (BERT) is critical (1st and 2nd rows vs 3--5th rows);

\begin{table}[t!] 
\centering
\caption{ Comparison of models under data-hungry environment. 
Logical pattern accuracy (P) and full logical form accuracy (LF) on test set of WikiSQL are shown. The errors are estimated by three independent experiments with different random seeds except \sqlovaglove\ where the error is estimated from two independent experiments.
} 
\label{tbl_rsset}
\footnotesize
\begin{threeparttable}
\begin{tabular}{lllll}
\toprule
Model  & Train set & Dev set & P (\%) & LF (\%) \\
\midrule
\ctof$^a$ & \rsset & \rssetd & - & $2.1 \pm 0.0$  \\ 
\sqlovaglove$^b$ & \rsset & \rssetd & $66.6 \pm 0.4$ & $17.6 \pm 0.3$  \\ 
\midrule
\sqlova$^b$ & \rsset  & \rssetd & $ 75.3 \pm 0.4$ & $ 45.1\pm 0.7$ \\
\ours\ w/o Quora  & \rsset  & \rssetd & $74.1 \pm 0.8$ & $49.1 \pm 0.9$ \\ 
\ours & \rsset  & \rssetd & $ 75.5 \pm 0.6$ & $50.0 \pm 0.6$ \\ 
\bottomrule
\end{tabular}
\begin{tablenotes}
\footnotesize
\item[a]{The source code is downloaded from https://github.com/donglixp/coarse2fine}
\item[b]{The source code is downloaded from https://github.com/naver/sqlova.}
\end{tablenotes}
\end{threeparttable}
\end{table}

It is of note that \ctof\ \cite{dongC2F2018} shows much lower accuracy compared to \sqlovaglove\ although both models use GLoVe \cite{pennington2014glove}. One possible explanation will be that \ctof\ first classify SQL patterns of \texttt{where} clause (sketch generation) while \sqlova\ generate SQL query via slot-filling approach. The classification involves abstraction of whole sentence and this process can be a data-hungry step.

\subsection{ Generalization test I: dependency on logical pattern distribution}
\begin{figure}[ht!] 
	\centering
	\includegraphics[width=0.3\textwidth]{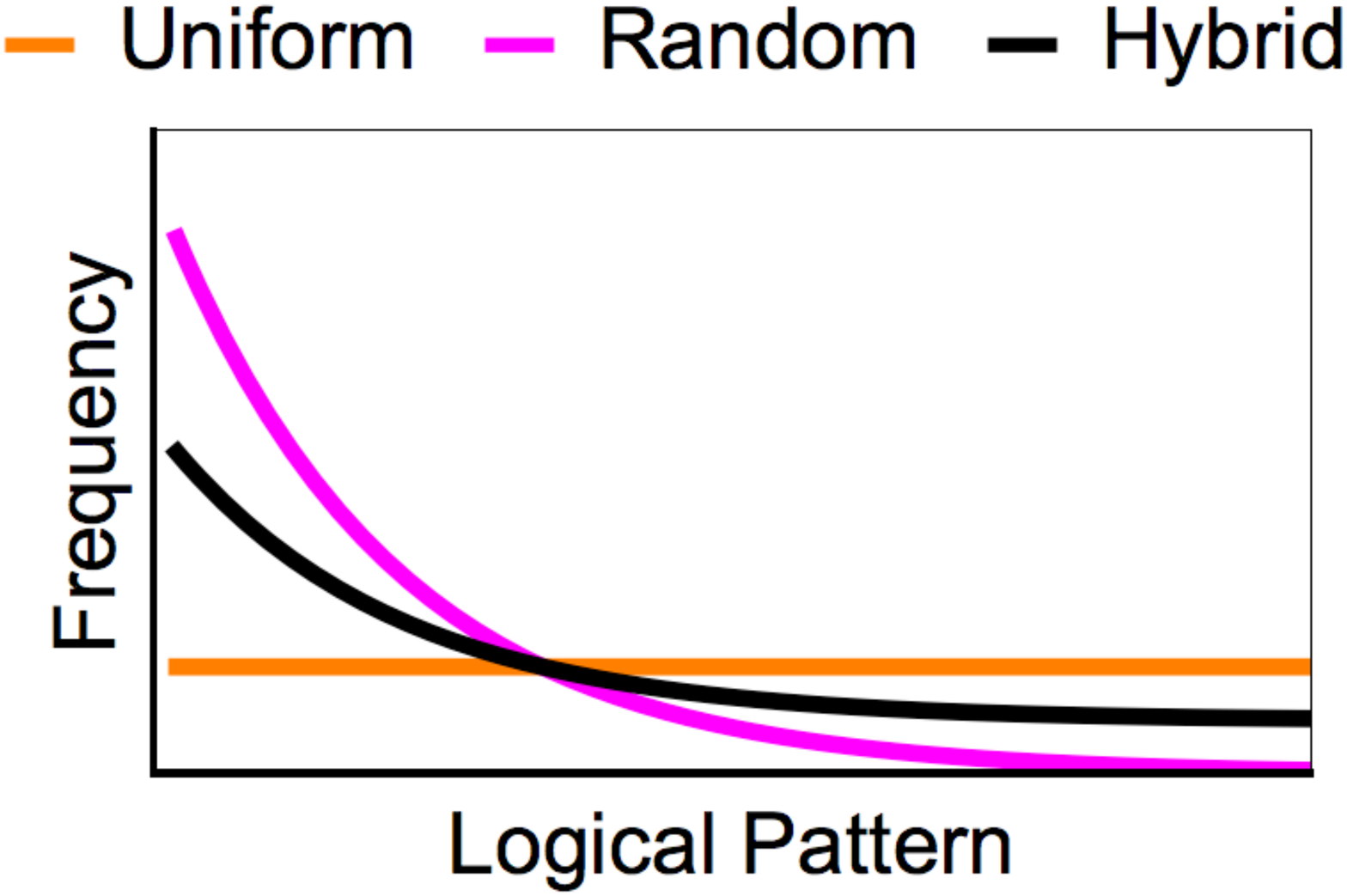}
	\caption{ The schematic plot of logical pattern distribution of three types of train sets: uniform set (orange), random set (magenta), and hybrid set  (black). In hybrid set, examples are distributed on logical patterns similar to random set but each logical pattern must include at least certain number of examples.
	}
	\label{fig_URH}
\end{figure}

When the size of train set is fixed, assigning more examples to frequently appearing logical patterns (in test environment) to the train set will increase the chance for correct SQL query generation as trained model would have a higher performance for frequent patterns (\rsset\ is constructed in this regard).
On the other hand, including diverse patterns in train set will help the model to distinguish similar patterns.
Considering these two aspects, we prepare additional two subsets \usset, and \hsset.
\usset\ consists of 850 uniformly distributed examples over 85 patterns whereas \ussetd\ consists of 320 uniformly distributed examples over 80 patterns.
\hsset\ is prepared by randomly sampling examples from top most frequent 85 logical patterns. Each pattern has approximately 128 times smaller number of examples compared to the full WikiSQL train set as in \rsset. In addition, all patterns are forced to have at least 7 examples for the diversity (Fig. \ref{fig_pattern}, and Table~\ref{tab:pattern_count}) resulting in total 897 examples. Only 85 patterns out of 210 patterns are considered because (1) 85 patterns occupy 98.6\% of full train set, and (2) only these patterns have at least 30 corresponding examples (Fig. \ref{fig_pattern}, Table~\ref{tab:pattern_count}). A dev set \hssetd\ is constructed similarly by extracting 223 examples from the WikiSQL dev set (Fig. \ref{fig_pattern}, Table~\ref{tab:pattern_count}).
The difference between three types of train sets are shown schematically in Fig. \ref{fig_URH} (orange: \usset, purple: \rsset, black: \hsset).

Table~\ref{tbl_usset_hsset} shows following important information: (1) \ours\ outperforms \sqlova\ again by +4.1\% LF in \usset\ (3rd and 5th rows of upper panel) and +4.0\% LF in \hsset\ (3rd and 5th rows of bottom panel); (2) the Quora pre-training improves model performance +5.9\% LF in \usset\ and by +0.5\% LF in \hsset (4th and 5th rows of each panel). 

\begin{table}[ht!] 
\centering
\caption{ Comparison of models with two additional train sets: \usset\ and \rsset.
} 
\label{tbl_usset_hsset}
\footnotesize
\begin{tabular}{lllll}
\toprule
Model  & Train set & Dev set & P (\%) & LF (\%) \\
\midrule
\sqlova & \usset  & \ussetd & $62.2 \pm 1.6$ & $33.8 \pm 1.2$ \\
\ours\ w/o Quora  & \usset  & \ussetd & $55.1 \pm 3.0$ & $32.0 \pm 1.7$ \\ 
\ours & \usset  & \ussetd & $65.7 \pm 0.4$ & $37.9 \pm 0.4$ \\ 
\midrule
\sqlova & \hsset  & \hssetd & $ 77.0 \pm 0.6$ & $ 45.9 \pm 0.4$ \\
\ours\ w/o Quora  & \hsset  & \hssetd & $ 78.5\pm 0.8$ & $ 49.4 \pm 1.2$ \\ 
\ours & \hsset  & \hssetd & $ 78.2\pm0.3  $& $ 49.9\pm 1.1 $ \\ 
\bottomrule
\end{tabular}
\end{table}
Both \ours\ and \sqlova\ show good performance when they are trained using either \rsset\ or \hsset (3rd and 5th columns of Table \ref{tbl_rsset}, \ref{tbl_usset_hsset}). In real service delivering scenario, the data distribution in test environment could vary with time. In regard of this, we prepare an additional test set \tusset\ by extracting 8 examples from top most frequent 81 logical patterns from the WikiSQL test. The resulting test set has completely different logical pattern distribution with the WikiSQL test set. 
The table \ref{tbl_tusset} shows that both models show best overall performance when they are trained with \hsset\ being remained robust to the change of test environment (4th columns). The result highlights the two important properties for train set to have: reflecting test environment (more examples for frequent logical patterns), and including diverse patterns.

\begin{table}[ht!] 
\centering
\caption{ Comparison of models with \tusset\ having uniform pattern distribution. The numbers in the table indicates LF of two models. The model with higher score in each condition is indicated by bold face.
} 
\label{tbl_tusset}
\footnotesize
\begin{tabular}{lccc}
\toprule
Model \& Test set  & \rsset & \usset & \hsset \\
\midrule
\sqlova, \tfset &$45.1 \pm 0.7$ & $33.8 \pm 1.2$& $45.9 \pm 0.4$ \\
\sqlova, \tusset &$\textbf{18.9} \pm 1.5$ & $32.3 \pm 1.3$& $31.7 \pm 1.3$ \\
\midrule
\ours, \tfset & $\textbf{50.0} \pm 0.6$ & $\textbf{37.9} \pm 0.4$ & $\textbf{49.9} \pm 1.1$\\
\ours, \tusset & $17.2 \pm 1.3$ & $ \textbf{39.2} \pm 1.2$ &  $\textbf{37.6} \pm 1.7$\\
\bottomrule
\end{tabular}
\end{table}

\subsection{Generalization test II: dependency on dataset size}

\begin{figure}[th!] 
	\centering
	\includegraphics[width=0.4\textwidth]{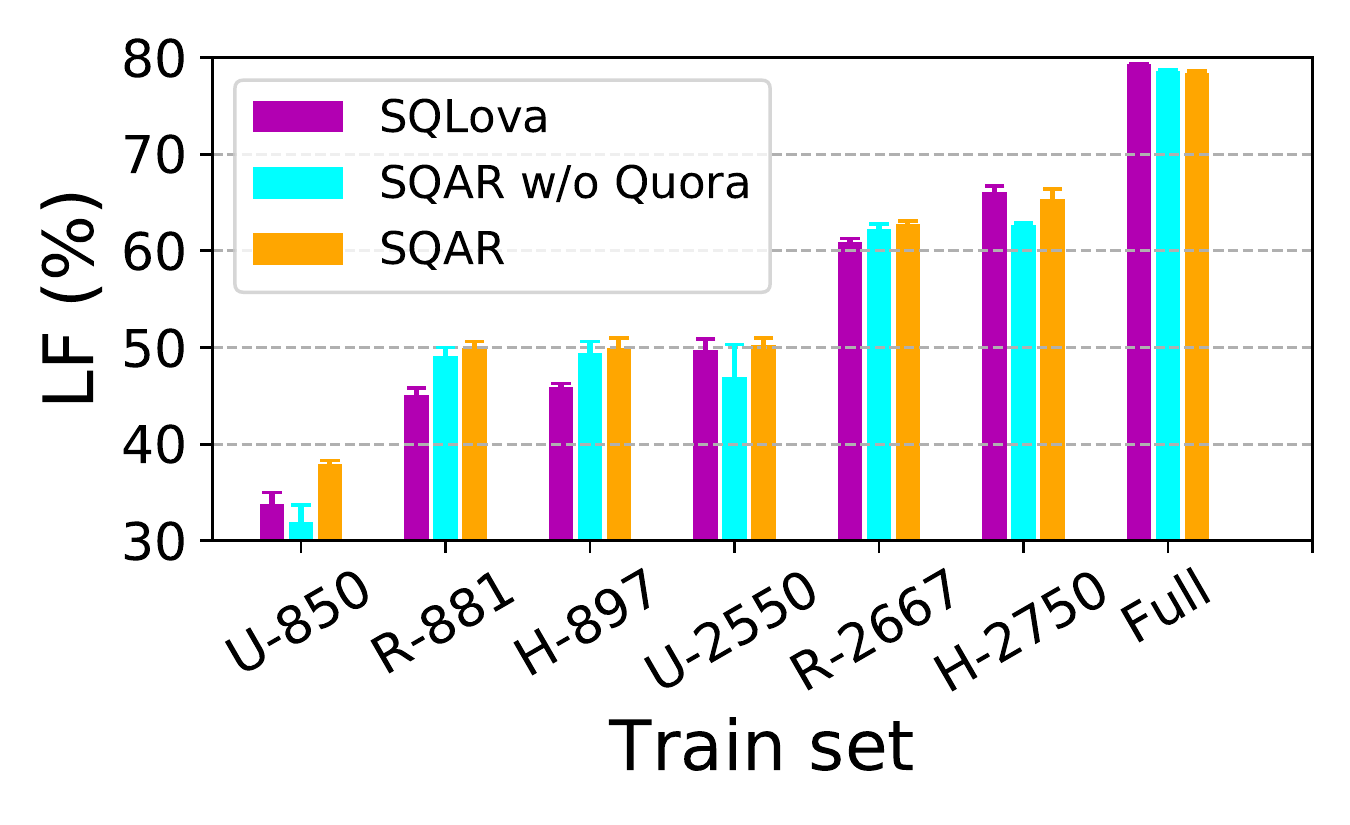}
	\caption{ Logical form accuracy of two models: \sqlova\ (magenta), \ours\ without Quora training (cyan), and \ours\ (orange) over various subsets (U-850: \usset, R-881: \rsset, H-897: \hsset, U-2550: \umset, R-2667: \rrmset, H-2750: \hhhhmset, Full: the WikiSQL train set)
	}
	\label{fig_LX_comp}
\end{figure}

To further test generality of our findings under change of train set size, we prepare three additional train sets: \umset, \rrmset, and \hhhhmset\ (Table~\ref{tab:pattern_count}).  
\umset\ consists of 2550 uniformly distributed examples over 85 patterns, \rrmset\ consists of 2667 examples randomly sampled from the WikiSQL train data, and \hhhhmset\ is larger version of \hsset\ in which each logical pattern includes at least 15 examples for 96 logical patterns (Table. \ref{tab:pattern_count}).
Table \ref{tbl_mset_fset} shows following information: (1) \ours\ shows marginally better performance than \sqlova\ showing +1.9\%, +0.5\%, and -0.7\% in LF when \rrmset, \umset, and \hhhhmset\ are used as the train sets (1st and 3rd rows of each panel); (2) Again, the pre-training using Quora paraphrasing datset increases LF by +0.5\%, +3.3\%, and +2.7\% in \rrmset, \umset, and \hhhhmset\ respectively (2nd and 3th rows of each panel); (3) Both \ours\ and \sqlova\ show best performance when they are trained over hybrid dataset. 
Observing that the performance gap between \ours\ and \sqlova\ becomes marginal as increasing the size of train set, we train both models using full WikiSQL train set. The result shows that again, there is only marginal difference between two models (\sqlova\ LF: $79.2 \pm 0.1$, \ours\ LF = $78.4 \pm 0.2$).
The overall results are summarized in Fig. \ref{fig_LX_comp}.

\begin{table*}[ht!] 
\centering
\caption{ 
Comparison of model with three WikiSQL train subsets: \rrmset, \umset\ and \hhhhmset).
}

\label{tbl_mset_fset}
\footnotesize
\begin{tabular*}{1.0\textwidth}{l@{\extracolsep{\fill}}lllll}
\toprule
Model & Train set & Dev set & P (\%) &  LF (\%) \\ 
\midrule
\sqlova & \rrmset & \rmsetd & $81.2 \pm 0.2$  & $60.9 \pm 0.4$ \\
\ours\ w/o  Quora & \rrmset & \rmsetd & $82.0 \pm 0.2$ & $62.3 \pm 0.5$ \\
\ours & \rrmset & \rmsetd & $81.4 \pm 0.5$ & $62.8 \pm 0.3$ \\ 
\midrule

\sqlova & \umset & \umsetd  & $68.2 \pm 1.6$  & $49.7 \pm 1.2$  \\ %
\ours\ w/o  Quora & \umset & \umsetd  & $66.2 \pm 4.5$  & $47.0 \pm 3.3$  \\ %
\ours & \umset & \umsetd  & $69.0 \pm 1.2$  & $50.3 \pm 0.7$ \\ %

\midrule    

\sqlova & \hhhhmset & \hmsetd  & $83.1 \pm 0.2$ & $66.1 \pm 0.6$  \\ 
\ours\ w/o  Quora & \hhhhmset & \hmsetd  & $82.2 \pm 0.2$ & $62.7 \pm 0.2 $ \\ 
\ours & \hhhhmset & \hmsetd  & $82.8 \pm 0.4$ & $65.4 \pm 1.0$  \\ 

\bottomrule
\end{tabular*}
\end{table*}


\subsection{ Generalization test III: parsing unseen logical forms} \label{sec:generalization}
In general, retrieval-based approach cannot handle new type of questions when corresponding logical patterns are not presented in the train set. However, unlike simple classification approach \cite{fineganDollak2018ACLmethodology}, \ours\ has interesting generalization ability originated from the use of query-to-query similarity in natural language space.
The train data in \ours\ has two roles: (1) supervision examples at training stage, and (2) a database to retrieve the logical pattern (a retrieval set) from which the most similar natural language query will be found during inference stage.
Once the model is trained, the second role can be improved by including more examples into the train set later. Particularly, by adding examples with new logical patterns, the model can handle questions with unseen logical patterns without re-training.

\begin{table}[t!] 
\centering
\caption{ 
Parsing unseen logical forms. \ours\ is trained by using \rsset\ and P and LF are measured while using a different set for query retrieval in the inference stage. R-881, H-897, H-2750, and Full stand for \rsset, \hsset, \hhhhmset, and \ufset\ respectively. 
R-capacity indicates the number of successfully retrieved logical pattern types whereas RG-capacity indicates that of successfully parsed logical pattern types.
}
\label{tbl_template_change}
\footnotesize
\begin{tabular}{l@{\extracolsep{\fill}}rllcccr}
\toprule
Model & Train set & Set for retrieval & P (\%) & LF (\%) & R-capacity & RG-capacity\\
\midrule
\ours & R-881 & R-881  & $75.5 \pm 0.6$ & $ 50.0 \pm 0.6$ & $57.5 \pm 2.2$ &  $47.3 \pm 0.4$\\
\ours & R-881 & R-881 + H-897  & $76.6 \pm 0.4$ & $50.6 \pm 0.5$  &  $79.3 \pm 1.9$ & $58.8 \pm 3.9$\\
\ours & R-881 & R-881 + H-2750  & $77.5 \pm 0.4$ & $50.7 \pm 0.6$  & $91.0 \pm 2.3$ & $67.0 \pm 1.7$\\
\ours & R-881 & Full  & $79.6 \pm 0.4$ & $51.7 \pm 0.5$ & $102 \pm 2$ &  $67.3 \pm 2.5$ \\
\midrule
\ours & R-881 & H-2750  & $77.2 \pm 0.5$ & $50.5 \pm 0.5$  & $92.0 \pm 2.1$ &  $67.5 \pm 2.2$\\
\bottomrule
\end{tabular}
\end{table}

To experimentally show this, we measured P and LF of \ours\ while changing the retrieval set during an inference stage (Table. \ref{tbl_template_change}). 
The train set is fixed to \rsset\ consisting of 67 logical patterns.
The result shows that upon addition of \hsset\ into the template set, which includes 18 more logical patterns compared to \rsset,
 P and LF increases by 1.1\% and 0.6\% respectively (2nd row of the table).
 Similar results are observed with \hhhhmset\ (+2.0\% in P and +0.7\% LF, 3rd row of the table) and with \tfset\ (+4.1\% in P and +1.7\% in LF, 4th row of the table).
To further show the power of using query-to-query similarity, we replaced the entire retrieval set from \rsset\ to \hhhhmset\ where only 43 examples are overlapped between them. Again, P and LF increase by 1.7\% and 0.5\% respectively (5th row of the table). 
To further confirm the addition of examples enables parsing of unseen logical patterns, we introduce two additional metrics: R-capacity and RG-capacity. R-capacity is defined by the number of successfully retrieved logical pattern types by \ours\ in the test set whereas RG-capacity indicates the number of successfully generated (retrieved and grounded) logical pattern types.
The table shows both R- and RG-capacities increases upon addition of examples into the retrieval set (5th and 6th columns). 
It should be emphasized that, during the training stage, \ours\ observed only 67 logical patterns.
Collectively, these results show that, \ours\ can be easily generalized to handle new logical patterns by simply adding new examples without re-training.
This also shows the possibility of transfer learning, even between semantic parsing tasks using different logical forms as intermediate logical patterns can be obtained from the natural language space.

\section{Conclusion}
We found that our retrieval-based model using query-to-query similarity can achieve high performance in WikiSQL semantic parsing task even when labeled data is scarce. We also found, pre-training using natural language paraphrasing data can help generation of logical forms in our query-similarity-based-retrieval approach. We also show that retrieval-based semantic parser can generate unseen logical forms during training stage.
Finally, we found careful design of data distribution is necessary for optimal performance of the model under data-scarce environment.



\begin{thebibliography}{24}
	\providecommand{\natexlab}[1]{#1}
	\providecommand{\url}[1]{\texttt{#1}}
	\expandafter\ifx\csname urlstyle\endcsname\relax
	\providecommand{\doi}[1]{doi: #1}\else
	\providecommand{\doi}{doi: \begingroup \urlstyle{rm}\Url}\fi
	
	\bibitem[Berant and Liang(2014)]{berant-liang-2014-semantic}
	Jonathan Berant and Percy Liang.
	\newblock Semantic parsing via paraphrasing.
	\newblock In \emph{Proceedings of the 52nd Annual Meeting of the Association
		for Computational Linguistics (Volume 1: Long Papers)}, pages 1415--1425,
	Baltimore, Maryland, June 2014. Association for Computational Linguistics.
	\newblock \doi{10.3115/v1/P14-1133}.
	\newblock URL \url{https://www.aclweb.org/anthology/P14-1133}.
	
	\bibitem[Devlin et~al.(2018)Devlin, Chang, Lee, and Toutanova]{devlinBERT2018}
	Jacob Devlin, Ming{-}Wei Chang, Kenton Lee, and Kristina Toutanova.
	\newblock {BERT:} pre-training of deep bidirectional transformers for language
	understanding.
	\newblock \emph{NAACL}, abs/1810.04805, 2018.
	\newblock URL \url{http://arxiv.org/abs/1810.04805}.
	
	\bibitem[Dong and Lapata(2016)]{dong2016Neural_semantic_parsing}
	Li~Dong and Mirella Lapata.
	\newblock Language to logical form with neural attention.
	\newblock In \emph{Proceedings of the 54th Annual Meeting of the Association
		for Computational Linguistics (Volume 1: Long Papers)}, pages 33--43, Berlin,
	Germany, August 2016. Association for Computational Linguistics.
	\newblock \doi{10.18653/v1/P16-1004}.
	\newblock URL \url{https://www.aclweb.org/anthology/P16-1004}.
	
	\bibitem[Dong and Lapata(2018)]{dongC2F2018}
	Li~Dong and Mirella Lapata.
	\newblock Coarse-to-fine decoding for neural semantic parsing.
	\newblock In \emph{Proceedings of the 56th Annual Meeting of the Association
		for Computational Linguistics (Volume 1: Long Papers)}, pages 731--742,
	Melbourne, Australia, July 2018. Association for Computational Linguistics.
	\newblock URL \url{https://www.aclweb.org/anthology/P18-1068}.
	
	\bibitem[Finegan-Dollak et~al.(2018)Finegan-Dollak, Kummerfeld, Zhang,
	Ramanathan, Sadasivam, Zhang, and Radev]{fineganDollak2018ACLmethodology}
	Catherine Finegan-Dollak, Jonathan~K. Kummerfeld, Li~Zhang, Karthik Ramanathan,
	Sesh Sadasivam, Rui Zhang, and Dragomir Radev.
	\newblock Improving text-to-sql evaluation methodology.
	\newblock In \emph{Proceedings of the 56th Annual Meeting of the Association
		for Computational Linguistics (Volume 1: Long Papers)}, pages 351--360.
	Association for Computational Linguistics, 2018.
	\newblock URL \url{http://aclweb.org/anthology/P18-1033}.
	
	\bibitem[He et~al.(2019)He, Mao, Chakrabarti, and Chen]{he2019xsql}
	Pengcheng He, Yi~Mao, Kaushik Chakrabarti, and Weizhu Chen.
	\newblock X-sql: Reinforce context into schema representation.
	\newblock Technical report, 2019.
	\newblock URL
	\url{https://www.microsoft.com/en-us/research/uploads/prod/2019/03/X_SQL-5c7db555d760f.pdf}.
	
	\bibitem[Hwang et~al.(2019)Hwang, Yim, Park, and Seo]{hwang2019sqlova}
	Wonseok Hwang, Jinyeong Yim, Seunghyun Park, and Minjoon Seo.
	\newblock A comprehensive exploration on wikisql with table-aware word
	contextualization.
	\newblock \emph{CoRR}, abs/1902.01069, 2019.
	\newblock URL \url{http://arxiv.org/abs/1902.01069}.
	
	\bibitem[Iyer et~al.(2017{\natexlab{a}})Iyer, Dandekar, and Csernai]{quoradata}
	Shankar Iyer, Nikhil Dandekar, and Kornél Csernai.
	\newblock First quora dataset release: Question pairs.
	\newblock 2017{\natexlab{a}}.
	\newblock URL \url{https://data.quora.com}.
	
	\bibitem[Iyer et~al.(2017{\natexlab{b}})Iyer, Konstas, Cheung, Krishnamurthy,
	and Zettlemoyer]{iyer2017UserInLoop}
	Srinivasan Iyer, Ioannis Konstas, Alvin Cheung, Jayant Krishnamurthy, and Luke
	Zettlemoyer.
	\newblock Learning a neural semantic parser from user feedback.
	\newblock In \emph{Proceedings of the 55th Annual Meeting of the Association
		for Computational Linguistics (Volume 1: Long Papers)}, pages 963--973,
	Vancouver, Canada, July 2017{\natexlab{b}}. Association for Computational
	Linguistics.
	\newblock \doi{10.18653/v1/P17-1089}.
	\newblock URL \url{https://www.aclweb.org/anthology/P17-1089}.
	
	\bibitem[Jia and Liang(2016)]{jia2016neuralSP_dataAug}
	Robin Jia and Percy Liang.
	\newblock Data recombination for neural semantic parsing.
	\newblock In \emph{Proceedings of the 54th Annual Meeting of the Association
		for Computational Linguistics (Volume 1: Long Papers)}, pages 12--22, Berlin,
	Germany, August 2016. Association for Computational Linguistics.
	\newblock \doi{10.18653/v1/P16-1002}.
	\newblock URL \url{https://www.aclweb.org/anthology/P16-1002}.
	
	\bibitem[Johnson et~al.(2017)Johnson, Douze, and J{\'e}gou]{johnson2017faiss}
	Jeff Johnson, Matthijs Douze, and Herv{\'e} J{\'e}gou.
	\newblock Billion-scale similarity search with gpus.
	\newblock \emph{arXiv preprint arXiv:1702.08734}, 2017.
	
	\bibitem[Manning et~al.(2014)Manning, Surdeanu, Bauer, Finkel, Bethard, and
	McClosky]{manningCoreNLP2014}
	Christopher~D. Manning, Mihai Surdeanu, John Bauer, Jenny Finkel, Steven~J.
	Bethard, and David McClosky.
	\newblock The {Stanford} {CoreNLP} natural language processing toolkit.
	\newblock In \emph{Association for Computational Linguistics (ACL) System
		Demonstrations}, pages 55--60, 2014.
	\newblock URL \url{http://www.aclweb.org/anthology/P/P14/P14-5010}.
	
	\bibitem[McCann et~al.(2018)McCann, Keskar, Xiong, and
	Socher]{mcCann2018decaNLP}
	Bryan McCann, Nitish~Shirish Keskar, Caiming Xiong, and Richard Socher.
	\newblock The natural language decathlon: Multitask learning as question
	answering.
	\newblock \emph{arXiv preprint arXiv:1806.08730}, 2018.
	
	\bibitem[Pennington et~al.(2014)Pennington, Socher, and
	Manning]{pennington2014glove}
	Jeffrey Pennington, Richard Socher, and Christopher~D. Manning.
	\newblock Glove: Global vectors for word representation.
	\newblock In \emph{Empirical Methods in Natural Language Processing (EMNLP)},
	pages 1532--1543, 2014.
	\newblock URL \url{http://www.aclweb.org/anthology/D14-1162}.
	
	\bibitem[Shi et~al.(2018)Shi, Tatwawadi, Chakrabarti, Mao, Polozov, and
	Chen]{shi2018IncSQL}
	Tianze Shi, Kedar Tatwawadi, Kaushik Chakrabarti, Yi~Mao, Oleksandr Polozov,
	and Weizhu Chen.
	\newblock Incsql: Training incremental text-to-sql parsers with
	non-deterministic oracles.
	\newblock \emph{CoRR}, abs/1809.05054, 2018.
	\newblock URL \url{http://arxiv.org/abs/1809.05054}.
	
	\bibitem[Vinyals et~al.(2015)Vinyals, Fortunato, and
	Jaitly]{vinyals2015NIPS_pointerNetwork}
	Oriol Vinyals, Meire Fortunato, and Navdeep Jaitly.
	\newblock Pointer networks.
	\newblock In C.~Cortes, N.~D. Lawrence, D.~D. Lee, M.~Sugiyama, and R.~Garnett,
	editors, \emph{Advances in Neural Information Processing Systems 28}, pages
	2692--2700. Curran Associates, Inc., 2015.
	\newblock URL \url{http://papers.nips.cc/paper/5866-pointer-networks.pdf}.
	
	\bibitem[Wang et~al.(2017)Wang, Brockschmidt, and Singh]{wang2017pointingOut}
	Chenglong Wang, Marc Brockschmidt, and Rishabh Singh.
	\newblock Pointing out {SQL} queries from text.
	\newblock Technical Report MSR-TR-2017-45, Microsoft, November 2017.
	\newblock URL
	\url{https://www.microsoft.com/en-us/research/publication/pointing-sql-queries-text/}.
	
	\bibitem[Wang et~al.(2018)Wang, Tian, Xiong, Wang, and Ku]{wang2018srllike}
	Wenlu Wang, Yingtao Tian, Hongyu Xiong, Haixun Wang, and Wei-Shinn Ku.
	\newblock A transfer-learnable natural language interface for databases.
	\newblock \emph{CoRR}, abs/1809.02649, 2018.
	
	\bibitem[Wu et~al.(2016)Wu, Schuster, Chen, Le, Norouzi, Macherey, Krikun, Cao,
	Gao, Macherey, Klingner, Shah, Johnson, Liu, Kaiser, Gouws, Kato, Kudo,
	Kazawa, Stevens, Kurian, Patil, Wang, Young, Smith, Riesa, Rudnick, Vinyals,
	Corrado, Hughes, and Dean]{wu2016word_piece}
	Yonghui Wu, Mike Schuster, Zhifeng Chen, Quoc~V. Le, Mohammad Norouzi, Wolfgang
	Macherey, Maxim Krikun, Yuan Cao, Qin Gao, Klaus Macherey, Jeff Klingner,
	Apurva Shah, Melvin Johnson, Xiaobing Liu, Lukasz Kaiser, Stephan Gouws,
	Yoshikiyo Kato, Taku Kudo, Hideto Kazawa, Keith Stevens, George Kurian,
	Nishant Patil, Wei Wang, Cliff Young, Jason Smith, Jason Riesa, Alex Rudnick,
	Oriol Vinyals, Greg Corrado, Macduff Hughes, and Jeffrey Dean.
	\newblock Google's neural machine translation system: Bridging the gap between
	human and machine translation.
	\newblock \emph{CoRR}, abs/1609.08144, 2016.
	\newblock URL \url{http://arxiv.org/abs/1609.08144}.
	
	\bibitem[Xiong and Sun(2018)]{xiong2018gan}
	Hongyu Xiong and Ruixiao Sun.
	\newblock Transferable natural language interface to structured queries aided
	by adversarial generation.
	\newblock \emph{CoRR}, abs/1812.01245, 2018.
	\newblock URL \url{http://arxiv.org/abs/1812.01245}.
	
	\bibitem[Xu et~al.(2017)Xu, Liu, and Song]{xuSQLNet2017}
	Xiaojun Xu, Chang Liu, and Dawn Song.
	\newblock Sqlnet: Generating structured queries from natural language without
	reinforcement learning.
	\newblock \emph{CoRR}, abs/1711.04436, 2017.
	\newblock URL \url{http://arxiv.org/abs/1711.04436}.
	
	\bibitem[Yin and Neubig(2018)]{yin2018TRANX}
	Pengcheng Yin and Graham Neubig.
	\newblock {TRANX}: A transition-based neural abstract syntax parser for
	semantic parsing and code generation.
	\newblock In \emph{Proceedings of the 2018 Conference on Empirical Methods in
		Natural Language Processing: System Demonstrations}, pages 7--12, Brussels,
	Belgium, November 2018. Association for Computational Linguistics.
	\newblock URL \url{https://www.aclweb.org/anthology/D18-2002}.
	
	\bibitem[Yu et~al.(2018)Yu, Li, Zhang, Zhang, and Radev]{yu2018TypeSQL}
	Tao Yu, Zifan Li, Zilin Zhang, Rui Zhang, and Dragomir Radev.
	\newblock {T}ype{SQL}: Knowledge-based type-aware neural text-to-{SQL}
	generation.
	\newblock In \emph{Proceedings of the 2018 Conference of the North {A}merican
		Chapter of the Association for Computational Linguistics: Human Language
		Technologies, Volume 2 (Short Papers)}, pages 588--594, New Orleans,
	Louisiana, June 2018. Association for Computational Linguistics.
	\newblock \doi{10.18653/v1/N18-2093}.
	\newblock URL \url{https://www.aclweb.org/anthology/N18-2093}.
	
	\bibitem[Zhong et~al.(2017)Zhong, Xiong, and Socher]{zhongSeq2SQL2017}
	Victor Zhong, Caiming Xiong, and Richard Socher.
	\newblock Seq2sql: Generating structured queries from natural language using
	reinforcement learning.
	\newblock \emph{CoRR}, abs/1709.00103, 2017.
	
\end{thebibliography}

\bibliographystyle{plainnat}

\newpage
\appendix
\onecolumn
\section{Appendix}
\setcounter{figure}{0}  
\setcounter{equation}{0}
\renewcommand{\thefigure}{A\arabic{figure}} 
\renewcommand{\theequation}{A\arabic{equation}}

\subsection{Experiments}
\subsubsection{Model training}
To train \ours, pre-trained BERT model (\texttt{BERT-Base-Uncased}\footnote{https://github.com/google-research/bert}) is loaded and fine-tuned using ADAM optimizer with learning rate of $2 \times 10^{-5}$ except the grounding module where the learning rate is set to $1 \times 10^{-3}$. The decay rates of ADAM optimizer are set to $\beta_1 = 0.9, \beta_2 = 0.999$.
Batch size is set to 12 for all experiment.
\sqlova\ is trained similarly using pre-trained BERT model (\texttt{BERT-Base-Uncased}). The learning rate is set to $1 \times 10^{-5}$ except NL2SQL layer which is trained with the learning rate $10^{-3}$. Batch size is set to 32 for all experiment.

Natural language utterance is first tokenized by using Standford CoreNLP~\cite{manningCoreNLP2014}.
Each token is further tokenized (into sub-word level) by WordPiece tokenizer~\cite{devlinBERT2018,wu2016word_piece}.
The headers of the tables and SQL vocabulary are tokenized by WordPiece tokenizer directly.
FAISS \cite{johnson2017faiss} is employed for the retrieval process.
The PyTorch version of BERT code\footnote{https://github.com/huggingface/pytorch-pre-trained-BERT} is used.
The model performance of \ctof\ was calculated by using the code\footnote{https://github.com/donglixp/coarse2fine} published by original authors \cite{dongC2F2018}. Our training of \ctof\ with the full WikiSQL train data results in $72 \pm 0.3$ logical form accuracy on WikiSQL test set.

All experiments were performed with WikiSQL ver. 1.1 \footnote{https://github.com/salesforce/WikiSQL}.
The model performance of \ours\, \sqlova\, and \ctof\ was measured by repeating three independent experiments in each condition with different random seeds. The errors are estimated by calculating standard deviation.
The performance of \sqlovaglove\ was measured from two independent experiment with different random seeds.
For the experiments with \usset, \rsset, \hsset, and \rmset, only single logical pattern is retrieved from the retriever due to the scarcity of examples per pattern. Otherwise 10 logical patterns are retrieved. The models are trained until the logical form accuracy is saturated waiting up to maximum 1000 epochs.

\subsubsection{Pre-training with Quora dataset}
To further pre-trained BERT-backbone used in \ours, we use Quora paraphrase detection dataset \cite{quoradata}. The dataset contains more than 405,000 question pairs with a corresponding binary indicator that represents whether two questions are a pair of paraphrase or not. The task setting is analogous to the retriever of \ours\ which detects the similarity of two given input NL queries and can be seen as fine-tuning in perspective of paraphrase detection task. During the training, two queries are given to the BERT model along with \texttt{[CLS]} and \texttt{[SEP]} tokens as in the original BERT training setting \cite{devlinBERT2018}. The output vector of \texttt{[CLS]} token was used for the binary classification to predict whether given two queries are a paraphrase pair or not. The model was trained until the classification accuracy converges using using ADAM optimizer.

\begin{figure*}[ht!] 
	\includegraphics[width=0.7\textwidth]{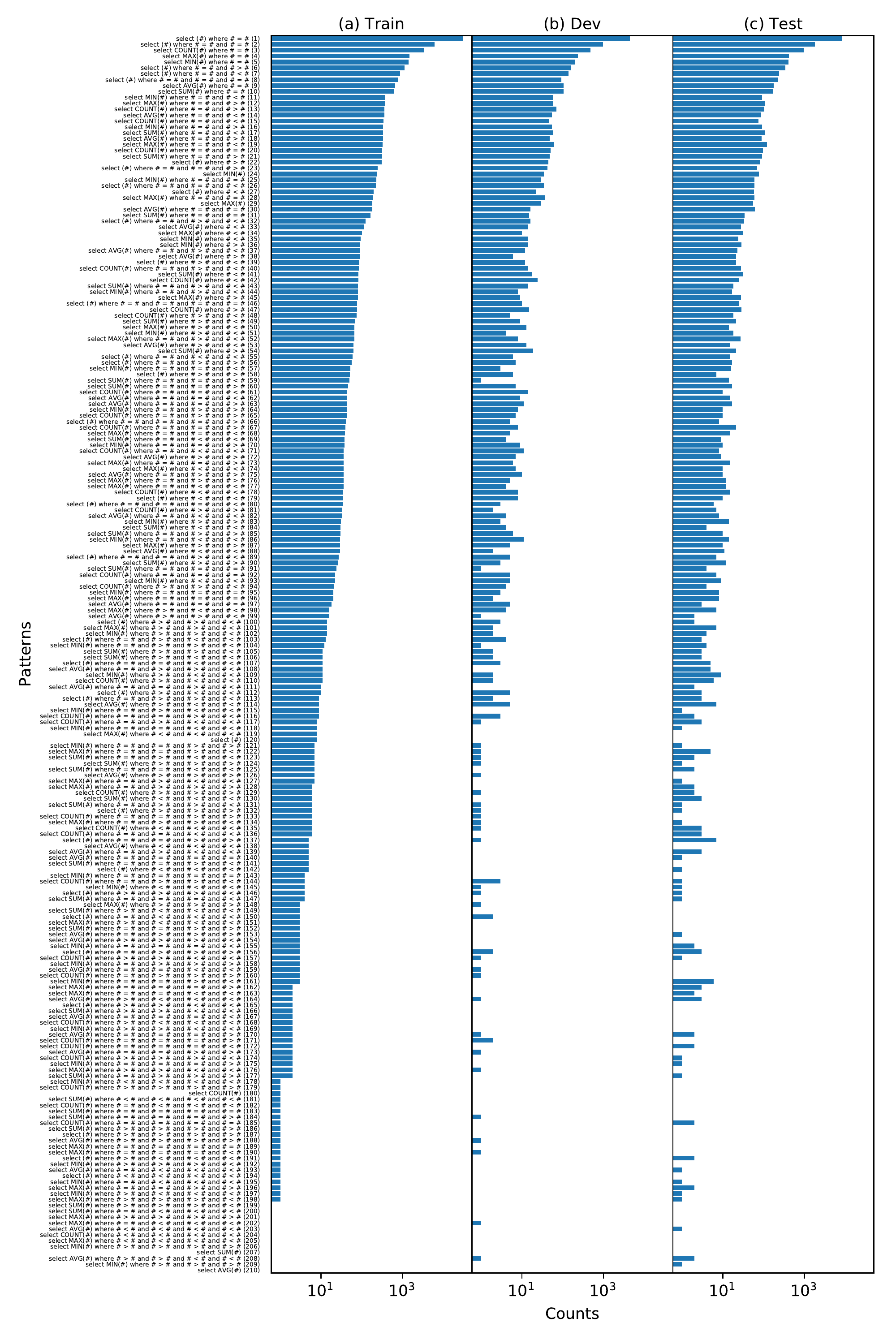}
	\caption{ SQL logical patterns and their frequency in the (a) train, (b) dev, and (c) test sets of WikiSQL. The index of each pattern is represented in the parentheses on the y-axis labels.
	}
	\label{fig_pattern}
\end{figure*}

\clearpage
\section{Supplementary tables}

\tiny
\begin{longtable}[ht!]{lrrrrrrrrrrrrrrr}
\caption{The count of SQL logical patterns in the WikiSQL subsets used in this paper. The subset names are denoted by following shorthand notations:
U-850 (\usset), R-881 (\rsset), H-897 (\hsset), U-2550 (\umset), R-2667 (\rrmset), H-2670 (\hmset), H-2750 (\hhhhmset), UD-320 (\ussetd), RD-132 (\rssetd), HD-223 (\hssetd), RD-527 (\rmsetd), HD-446 (\hmsetd)
}\\
\label{tab:pattern_count}\\
\toprule
\multicolumn{1}{p{0.5cm}}{\centering Pattern \\ index} & 
\multirow{2}{*}{Train} & \multirow{2}{*}{Dev} & \multirow{2}{*}{Test} & \multirow{2}{*}{U-850} & \multirow{2}{*}{R-881} & \multirow{2}{*}{H-897} & \multirow{2}{*}{U-2550} & \multirow{2}{*}{R-2667} & \multirow{2}{*}{H-2670} & \multirow{2}{*}{H-2750} & \multirow{2}{*}{UD-320} & \multirow{2}{*}{RD-132} & \multirow{2}{*}{HD-223} & \multirow{2}{*}{RD-527} & \multirow{2}{*}{HD-446} \\
\midrule
1 & 30128 & 4419 & 8269 & 10 & 484 & 236 & 30 & 1425 & 471 & 942 & 4 & 84 & 35 & 305 & 70 \\
2 & 6050 & 964 & 1798 & 10 & 95 & 48 & 30 & 263 & 95 & 190 & 4 & 10 & 8 & 52 & 16 \\
3 & 3430 & 480 & 972 & 10 & 61 & 27 & 30 & 159 & 54 & 108 & 4 & 4 & 4 & 23 & 8 \\
4 & 1476 & 235 & 419 & 10 & 23 & 12 & 30 & 84 & 25 & 47 & 4 & 5 & 2 & 19 & 4 \\
5 & 1412 & 202 & 413 & 10 & 22 & 12 & 30 & 65 & 25 & 45 & 4 & 3 & 2 & 7 & 4 \\
6 & 1116 & 158 & 339 & 10 & 14 & 9 & 30 & 53 & 25 & 35 & 4 & 3 & 2 & 15 & 4 \\
7 & 858 & 138 & 240 & 10 & 14 & 7 & 30 & 55 & 25 & 27 & 4 & 4 & 2 & 10 & 4 \\
8 & 794 & 92 & 229 & 10 & 13 & 7 & 30 & 30 & 25 & 25 & 4 & 1 & 2 & 3 & 4 \\
9 & 653 & 104 & 178 & 10 & 9 & 7 & 30 & 27 & 25 & 21 & 4 & 2 & 2 & 8 & 4 \\
10 & 619 & 105 & 173 & 10 & 11 & 7 & 30 & 30 & 25 & 20 & 4 & 0 & 2 & 3 & 4 \\
\midrule
11 & 374 & 57 & 91 & 10 & 8 & 7 & 30 & 25 & 25 & 15 & 4 & 2 & 2 & 6 & 4 \\
12 & 369 & 59 & 107 & 10 & 5 & 7 & 30 & 23 & 25 & 15 & 4 & 0 & 2 & 5 & 4 \\
13 & 360 & 70 & 103 & 10 & 7 & 7 & 30 & 21 & 25 & 15 & 4 & 0 & 2 & 3 & 4 \\
14 & 357 & 55 & 88 & 10 & 9 & 7 & 30 & 17 & 25 & 15 & 4 & 1 & 2 & 2 & 4 \\
15 & 341 & 46 & 75 & 10 & 7 & 7 & 30 & 11 & 25 & 15 & 4 & 1 & 2 & 2 & 4 \\
16 & 334 & 55 & 92 & 10 & 5 & 7 & 30 & 17 & 25 & 15 & 4 & 0 & 2 & 0 & 4 \\
17 & 330 & 59 & 109 & 10 & 3 & 7 & 30 & 16 & 25 & 15 & 4 & 0 & 2 & 1 & 4 \\
18 & 329 & 48 & 89 & 10 & 3 & 7 & 30 & 19 & 25 & 15 & 4 & 0 & 2 & 1 & 4 \\
19 & 324 & 62 & 120 & 10 & 7 & 7 & 30 & 21 & 25 & 15 & 4 & 0 & 2 & 4 & 4 \\
20 & 319 & 50 & 97 & 10 & 2 & 7 & 30 & 9 & 25 & 15 & 4 & 1 & 2 & 5 & 4 \\
\midrule
21 & 312 & 48 & 93 & 10 & 5 & 7 & 30 & 12 & 25 & 15 & 4 & 0 & 2 & 0 & 4 \\
22 & 309 & 45 & 83 & 10 & 2 & 7 & 30 & 15 & 25 & 15 & 4 & 0 & 2 & 0 & 4 \\
23 & 247 & 42 & 69 & 10 & 4 & 7 & 30 & 11 & 25 & 15 & 4 & 1 & 2 & 2 & 4 \\
24 & 234 & 35 & 76 & 10 & 3 & 7 & 30 & 10 & 25 & 15 & 4 & 1 & 2 & 1 & 4 \\
25 & 225 & 30 & 59 & 10 & 3 & 7 & 30 & 12 & 25 & 15 & 4 & 1 & 2 & 3 & 4 \\
26 & 222 & 35 & 59 & 10 & 2 & 7 & 30 & 9 & 25 & 15 & 4 & 0 & 2 & 0 & 4 \\
27 & 197 & 22 & 58 & 10 & 2 & 7 & 30 & 10 & 25 & 15 & 4 & 1 & 2 & 2 & 4 \\
28 & 188 & 36 & 59 & 10 & 3 & 7 & 30 & 7 & 25 & 15 & 4 & 0 & 2 & 0 & 4 \\
29 & 180 & 29 & 55 & 10 & 2 & 7 & 30 & 9 & 25 & 15 & 4 & 1 & 2 & 1 & 4 \\
30 & 179 & 16 & 61 & 10 & 2 & 7 & 30 & 12 & 25 & 15 & 4 & 0 & 2 & 0 & 4 \\
\midrule
31 & 165 & 15 & 34 & 10 & 0 & 7 & 30 & 6 & 25 & 15 & 4 & 0 & 2 & 2 & 4 \\
32 & 123 & 16 & 33 & 10 & 1 & 7 & 30 & 6 & 25 & 15 & 4 & 0 & 2 & 2 & 4 \\
33 & 116 & 14 & 28 & 10 & 1 & 7 & 30 & 7 & 25 & 15 & 4 & 0 & 2 & 0 & 4 \\
34 & 100 & 10 & 31 & 10 & 1 & 7 & 30 & 3 & 25 & 15 & 4 & 0 & 2 & 0 & 4 \\
35 & 93 & 14 & 24 & 10 & 2 & 7 & 30 & 6 & 25 & 15 & 4 & 0 & 2 & 2 & 4 \\
36 & 91 & 14 & 29 & 10 & 2 & 7 & 30 & 4 & 25 & 15 & 4 & 0 & 2 & 1 & 4 \\
37 & 89 & 12 & 23 & 10 & 1 & 7 & 30 & 5 & 25 & 15 & 4 & 0 & 2 & 0 & 4 \\
38 & 89 & 6 & 21 & 10 & 0 & 7 & 30 & 4 & 25 & 15 & 4 & 0 & 2 & 0 & 4 \\
39 & 86 & 12 & 21 & 10 & 2 & 7 & 30 & 7 & 25 & 15 & 4 & 0 & 2 & 0 & 4 \\
40 & 84 & 14 & 28 & 10 & 2 & 7 & 30 & 6 & 25 & 15 & 4 & 1 & 2 & 1 & 4 \\
\midrule
41 & 83 & 18 & 31 & 10 & 1 & 7 & 30 & 2 & 25 & 15 & 4 & 0 & 2 & 2 & 4 \\
42 & 82 & 24 & 25 & 10 & 3 & 7 & 30 & 3 & 25 & 15 & 4 & 0 & 2 & 1 & 4 \\
43 & 81 & 14 & 18 & 10 & 1 & 7 & 30 & 6 & 25 & 15 & 4 & 0 & 2 & 0 & 4 \\
44 & 81 & 8 & 17 & 10 & 2 & 7 & 30 & 3 & 25 & 15 & 4 & 1 & 2 & 1 & 4 \\
45 & 80 & 9 & 28 & 10 & 0 & 7 & 30 & 5 & 25 & 15 & 4 & 0 & 2 & 1 & 4 \\
46 & 77 & 10 & 25 & 10 & 1 & 7 & 30 & 3 & 25 & 15 & 4 & 2 & 2 & 2 & 4 \\
47 & 76 & 15 & 29 & 10 & 0 & 7 & 30 & 2 & 25 & 15 & 4 & 0 & 2 & 0 & 4 \\
48 & 74 & 5 & 18 & 10 & 0 & 7 & 30 & 3 & 25 & 15 & 4 & 0 & 2 & 0 & 4 \\
49 & 68 & 9 & 21 & 10 & 2 & 7 & 30 & 2 & 25 & 15 & 4 & 0 & 2 & 0 & 4 \\
50 & 66 & 13 & 14 & 10 & 1 & 7 & 30 & 3 & 25 & 15 & 4 & 0 & 2 & 0 & 4 \\
\midrule
51 & 66 & 4 & 18 & 10 & 2 & 7 & 30 & 4 & 25 & 15 & 4 & 0 & 2 & 0 & 4 \\
52 & 66 & 8 & 27 & 10 & 0 & 7 & 30 & 5 & 25 & 15 & 4 & 0 & 2 & 0 & 4 \\
53 & 63 & 13 & 15 & 10 & 2 & 7 & 30 & 5 & 25 & 15 & 4 & 0 & 2 & 0 & 4 \\
54 & 62 & 19 & 21 & 10 & 2 & 7 & 30 & 4 & 25 & 15 & 4 & 0 & 2 & 1 & 4 \\
55 & 59 & 6 & 15 & 10 & 0 & 7 & 30 & 4 & 25 & 15 & 4 & 0 & 2 & 1 & 4 \\
56 & 56 & 7 & 17 & 10 & 1 & 7 & 30 & 1 & 25 & 15 & 4 & 0 & 2 & 0 & 4 \\
57 & 52 & 3 & 16 & 10 & 0 & 7 & 30 & 2 & 25 & 15 & 0 & 0 & 0 & 1 & 0 \\
58 & 51 & 6 & 7 & 10 & 1 & 7 & 30 & 1 & 25 & 15 & 4 & 0 & 2 & 0 & 4 \\
59 & 50 & 1 & 14 & 10 & 0 & 7 & 30 & 0 & 25 & 15 & 0 & 0 & 0 & 0 & 0 \\
60 & 46 & 7 & 17 & 10 & 0 & 7 & 30 & 2 & 25 & 15 & 4 & 0 & 2 & 0 & 4 \\
\midrule
61 & 44 & 14 & 10 & 10 & 1 & 7 & 30 & 3 & 25 & 15 & 4 & 0 & 2 & 1 & 4 \\
62 & 44 & 9 & 15 & 10 & 0 & 7 & 30 & 3 & 25 & 15 & 4 & 0 & 2 & 2 & 4 \\
63 & 43 & 11 & 17 & 10 & 0 & 7 & 30 & 3 & 25 & 15 & 4 & 0 & 2 & 1 & 4 \\
64 & 43 & 8 & 10 & 10 & 1 & 7 & 30 & 4 & 25 & 15 & 4 & 0 & 2 & 0 & 4 \\
65 & 43 & 7 & 10 & 10 & 0 & 7 & 30 & 1 & 25 & 15 & 4 & 0 & 2 & 1 & 4 \\
66 & 41 & 5 & 8 & 10 & 0 & 7 & 30 & 2 & 25 & 15 & 4 & 0 & 2 & 2 & 4 \\
67 & 40 & 8 & 21 & 10 & 2 & 7 & 30 & 1 & 25 & 15 & 4 & 0 & 2 & 1 & 4 \\
68 & 39 & 5 & 15 & 10 & 0 & 7 & 30 & 3 & 25 & 15 & 4 & 0 & 2 & 0 & 4 \\
69 & 38 & 4 & 9 & 10 & 1 & 7 & 30 & 3 & 25 & 15 & 4 & 0 & 2 & 0 & 4 \\
70 & 38 & 9 & 10 & 10 & 0 & 7 & 30 & 4 & 25 & 15 & 4 & 0 & 2 & 1 & 4 \\
\midrule
71 & 36 & 11 & 8 & 10 & 0 & 7 & 30 & 1 & 25 & 15 & 4 & 0 & 2 & 3 & 4 \\
72 & 36 & 7 & 9 & 10 & 0 & 7 & 30 & 0 & 25 & 15 & 4 & 0 & 2 & 0 & 4 \\
73 & 36 & 6 & 15 & 10 & 0 & 7 & 30 & 1 & 25 & 15 & 4 & 2 & 2 & 2 & 4 \\
74 & 36 & 7 & 10 & 10 & 0 & 7 & 30 & 0 & 25 & 15 & 4 & 0 & 2 & 1 & 4 \\
75 & 36 & 10 & 10 & 10 & 0 & 7 & 30 & 2 & 25 & 15 & 4 & 0 & 2 & 2 & 4 \\
76 & 36 & 5 & 12 & 10 & 1 & 7 & 30 & 2 & 25 & 15 & 4 & 0 & 2 & 0 & 4 \\
77 & 36 & 4 & 12 & 10 & 0 & 7 & 30 & 1 & 25 & 15 & 4 & 0 & 2 & 0 & 4 \\
78 & 35 & 8 & 15 & 10 & 1 & 7 & 30 & 2 & 25 & 15 & 4 & 0 & 2 & 0 & 4 \\
79 & 35 & 8 & 10 & 10 & 1 & 7 & 30 & 2 & 25 & 15 & 4 & 0 & 2 & 0 & 4 \\
80 & 34 & 3 & 6 & 10 & 0 & 7 & 30 & 2 & 25 & 15 & 0 & 0 & 0 & 0 & 0 \\
\midrule
81 & 33 & 2 & 7 & 10 & 0 & 7 & 30 & 2 & 25 & 15 & 0 & 0 & 0 & 0 & 0 \\
82 & 33 & 4 & 8 & 10 & 0 & 7 & 30 & 2 & 25 & 15 & 4 & 0 & 2 & 0 & 4 \\
83 & 31 & 3 & 14 & 10 & 2 & 7 & 30 & 1 & 25 & 15 & 0 & 0 & 0 & 0 & 0 \\
84 & 30 & 4 & 4 & 10 & 0 & 7 & 30 & 2 & 25 & 15 & 4 & 0 & 2 & 0 & 4 \\
85 & 30 & 6 & 10 & 10 & 0 & 7 & 30 & 2 & 25 & 15 & 4 & 0 & 2 & 0 & 4 \\
86 & 29 & 11 & 14 & 0 & 0 & 0 & 0 & 2 & 0 & 15 & 0 & 0 & 2 & 2 & 4 \\
87 & 29 & 5 & 10 & 0 & 0 & 0 & 0 & 0 & 0 & 15 & 0 & 0 & 2 & 0 & 4 \\
88 & 29 & 2 & 11 & 0 & 0 & 0 & 0 & 0 & 0 & 15 & 0 & 0 & 0 & 0 & 0 \\
89 & 27 & 5 & 7 & 0 & 2 & 0 & 0 & 1 & 0 & 15 & 0 & 0 & 2 & 0 & 4 \\
90 & 26 & 3 & 12 & 0 & 0 & 0 & 0 & 0 & 0 & 15 & 0 & 0 & 0 & 0 & 0 \\
\midrule
91 & 24 & 1 & 4 & 0 & 0 & 0 & 0 & 0 & 0 & 15 & 0 & 0 & 0 & 0 & 0 \\
92 & 22 & 5 & 7 & 0 & 1 & 0 & 0 & 1 & 0 & 15 & 0 & 0 & 2 & 0 & 4 \\
93 & 22 & 5 & 9 & 0 & 0 & 0 & 0 & 2 & 0 & 15 & 0 & 0 & 2 & 0 & 4 \\
94 & 21 & 4 & 4 & 0 & 0 & 0 & 0 & 1 & 0 & 15 & 0 & 0 & 2 & 0 & 4 \\
95 & 20 & 3 & 8 & 0 & 0 & 0 & 0 & 1 & 0 & 15 & 0 & 0 & 0 & 0 & 0 \\
96 & 20 & 2 & 8 & 0 & 0 & 0 & 0 & 0 & 0 & 15 & 0 & 0 & 0 & 1 & 0 \\
97 & 18 & 5 & 3 & 0 & 0 & 0 & 0 & 0 & 0 & 0 & 0 & 0 & 2 & 0 & 4 \\
98 & 16 & 4 & 7 & 0 & 1 & 0 & 0 & 1 & 0 & 0 & 0 & 0 & 2 & 0 & 4 \\
99 & 16 & 1 & 2 & 0 & 0 & 0 & 0 & 0 & 0 & 0 & 0 & 0 & 0 & 0 & 0 \\
100 & 14 & 3 & 2 & 0 & 0 & 0 & 0 & 1 & 0 & 0 & 0 & 0 & 0 & 0 & 0 \\
\midrule
101 & 14 & 2 & 7 & 0 & 0 & 0 & 0 & 0 & 0 & 0 & 0 & 0 & 0 & 0 & 0 \\
102 & 14 & 2 & 4 & 0 & 0 & 0 & 0 & 0 & 0 & 0 & 0 & 0 & 0 & 0 & 0 \\
103 & 13 & 4 & 3 & 0 & 0 & 0 & 0 & 1 & 0 & 0 & 0 & 0 & 2 & 0 & 4 \\
104 & 12 & 1 & 4 & 0 & 0 & 0 & 0 & 1 & 0 & 0 & 0 & 0 & 0 & 0 & 0 \\
105 & 11 & 2 & 3 & 0 & 0 & 0 & 0 & 1 & 0 & 0 & 0 & 0 & 0 & 0 & 0 \\
106 & 11 & 2 & 3 & 0 & 0 & 0 & 0 & 0 & 0 & 0 & 0 & 0 & 0 & 0 & 0 \\
107 & 11 & 3 & 5 & 0 & 0 & 0 & 0 & 1 & 0 & 0 & 0 & 0 & 0 & 1 & 0 \\
108 & 11 & 0 & 5 & 0 & 0 & 0 & 0 & 0 & 0 & 0 & 0 & 0 & 0 & 0 & 0 \\
109 & 11 & 2 & 9 & 0 & 0 & 0 & 0 & 1 & 0 & 0 & 0 & 0 & 0 & 0 & 0 \\
110 & 11 & 2 & 6 & 0 & 1 & 0 & 0 & 0 & 0 & 0 & 0 & 0 & 0 & 0 & 0 \\
\midrule
111 & 10 & 0 & 2 & 0 & 0 & 0 & 0 & 1 & 0 & 0 & 0 & 0 & 0 & 0 & 0 \\
112 & 10 & 5 & 3 & 0 & 0 & 0 & 0 & 1 & 0 & 0 & 0 & 0 & 2 & 0 & 4 \\
113 & 9 & 2 & 3 & 0 & 0 & 0 & 0 & 0 & 0 & 0 & 0 & 0 & 0 & 0 & 0 \\
114 & 9 & 5 & 7 & 0 & 0 & 0 & 0 & 0 & 0 & 0 & 0 & 0 & 2 & 1 & 4 \\
115 & 9 & 0 & 1 & 0 & 0 & 0 & 0 & 0 & 0 & 0 & 0 & 0 & 0 & 0 & 0 \\
116 & 9 & 3 & 2 & 0 & 0 & 0 & 0 & 0 & 0 & 0 & 0 & 0 & 0 & 0 & 0 \\
117 & 8 & 1 & 3 & 0 & 0 & 0 & 0 & 0 & 0 & 0 & 0 & 0 & 0 & 0 & 0 \\
118 & 8 & 0 & 1 & 0 & 0 & 0 & 0 & 0 & 0 & 0 & 0 & 0 & 0 & 0 & 0 \\
119 & 8 & 0 & 0 & 0 & 0 & 0 & 0 & 1 & 0 & 0 & 0 & 0 & 0 & 0 & 0 \\
120 & 8 & 0 & 0 & 0 & 0 & 0 & 0 & 0 & 0 & 0 & 0 & 0 & 0 & 0 & 0 \\
\midrule
121 & 7 & 1 & 1 & 0 & 0 & 0 & 0 & 1 & 0 & 0 & 0 & 0 & 0 & 0 & 0 \\
122 & 7 & 1 & 5 & 0 & 0 & 0 & 0 & 0 & 0 & 0 & 0 & 0 & 0 & 0 & 0 \\
123 & 7 & 1 & 2 & 0 & 0 & 0 & 0 & 2 & 0 & 0 & 0 & 0 & 0 & 0 & 0 \\
124 & 7 & 1 & 1 & 0 & 0 & 0 & 0 & 1 & 0 & 0 & 0 & 0 & 0 & 0 & 0 \\
125 & 7 & 0 & 2 & 0 & 0 & 0 & 0 & 1 & 0 & 0 & 0 & 0 & 0 & 0 & 0 \\
126 & 7 & 1 & 0 & 0 & 1 & 0 & 0 & 1 & 0 & 0 & 0 & 0 & 0 & 0 & 0 \\
127 & 7 & 0 & 1 & 0 & 0 & 0 & 0 & 2 & 0 & 0 & 0 & 0 & 0 & 0 & 0 \\
128 & 6 & 0 & 2 & 0 & 0 & 0 & 0 & 1 & 0 & 0 & 0 & 0 & 0 & 0 & 0 \\
129 & 6 & 1 & 2 & 0 & 0 & 0 & 0 & 1 & 0 & 0 & 0 & 0 & 0 & 0 & 0 \\
130 & 6 & 0 & 3 & 0 & 0 & 0 & 0 & 0 & 0 & 0 & 0 & 0 & 0 & 0 & 0 \\
\midrule
131 & 6 & 1 & 1 & 0 & 0 & 0 & 0 & 1 & 0 & 0 & 0 & 0 & 0 & 0 & 0 \\
132 & 6 & 1 & 1 & 0 & 1 & 0 & 0 & 0 & 0 & 0 & 0 & 0 & 0 & 0 & 0 \\
133 & 6 & 1 & 0 & 0 & 0 & 0 & 0 & 0 & 0 & 0 & 0 & 0 & 0 & 0 & 0 \\
134 & 6 & 1 & 1 & 0 & 1 & 0 & 0 & 0 & 0 & 0 & 0 & 0 & 0 & 1 & 0 \\
135 & 6 & 1 & 3 & 0 & 0 & 0 & 0 & 0 & 0 & 0 & 0 & 0 & 0 & 0 & 0 \\
136 & 6 & 0 & 3 & 0 & 0 & 0 & 0 & 1 & 0 & 0 & 0 & 0 & 0 & 0 & 0 \\
137 & 5 & 1 & 7 & 0 & 0 & 0 & 0 & 0 & 0 & 0 & 0 & 0 & 0 & 0 & 0 \\
138 & 5 & 0 & 0 & 0 & 0 & 0 & 0 & 1 & 0 & 0 & 0 & 0 & 0 & 0 & 0 \\
139 & 5 & 0 & 3 & 0 & 0 & 0 & 0 & 0 & 0 & 0 & 0 & 0 & 0 & 0 & 0 \\
140 & 5 & 0 & 1 & 0 & 0 & 0 & 0 & 0 & 0 & 0 & 0 & 0 & 0 & 0 & 0 \\
\midrule
141 & 5 & 0 & 0 & 0 & 1 & 0 & 0 & 0 & 0 & 0 & 0 & 0 & 0 & 0 & 0 \\
142 & 5 & 0 & 1 & 0 & 0 & 0 & 0 & 0 & 0 & 0 & 0 & 0 & 0 & 0 & 0 \\
143 & 4 & 0 & 0 & 0 & 0 & 0 & 0 & 0 & 0 & 0 & 0 & 0 & 0 & 0 & 0 \\
144 & 4 & 3 & 1 & 0 & 0 & 0 & 0 & 1 & 0 & 0 & 0 & 0 & 0 & 1 & 0 \\
145 & 4 & 1 & 1 & 0 & 0 & 0 & 0 & 0 & 0 & 0 & 0 & 0 & 0 & 0 & 0 \\
146 & 4 & 1 & 1 & 0 & 0 & 0 & 0 & 0 & 0 & 0 & 0 & 0 & 0 & 0 & 0 \\
147 & 4 & 0 & 1 & 0 & 0 & 0 & 0 & 0 & 0 & 0 & 0 & 0 & 0 & 0 & 0 \\
148 & 3 & 1 & 0 & 0 & 0 & 0 & 0 & 0 & 0 & 0 & 0 & 0 & 0 & 0 & 0 \\
149 & 3 & 0 & 0 & 0 & 0 & 0 & 0 & 0 & 0 & 0 & 0 & 0 & 0 & 0 & 0 \\
150 & 3 & 2 & 0 & 0 & 0 & 0 & 0 & 0 & 0 & 0 & 0 & 0 & 0 & 1 & 0 \\
\midrule
151 & 3 & 0 & 0 & 0 & 0 & 0 & 0 & 0 & 0 & 0 & 0 & 0 & 0 & 0 & 0 \\
152 & 3 & 0 & 0 & 0 & 0 & 0 & 0 & 0 & 0 & 0 & 0 & 0 & 0 & 0 & 0 \\
153 & 3 & 0 & 1 & 0 & 0 & 0 & 0 & 0 & 0 & 0 & 0 & 0 & 0 & 0 & 0 \\
154 & 3 & 0 & 0 & 0 & 0 & 0 & 0 & 0 & 0 & 0 & 0 & 0 & 0 & 0 & 0 \\
155 & 3 & 0 & 2 & 0 & 0 & 0 & 0 & 0 & 0 & 0 & 0 & 0 & 0 & 0 & 0 \\
156 & 3 & 2 & 3 & 0 & 0 & 0 & 0 & 1 & 0 & 0 & 0 & 0 & 0 & 0 & 0 \\
157 & 3 & 1 & 1 & 0 & 0 & 0 & 0 & 0 & 0 & 0 & 0 & 0 & 0 & 0 & 0 \\
158 & 3 & 0 & 0 & 0 & 0 & 0 & 0 & 0 & 0 & 0 & 0 & 0 & 0 & 0 & 0 \\
159 & 3 & 1 & 0 & 0 & 0 & 0 & 0 & 1 & 0 & 0 & 0 & 0 & 0 & 0 & 0 \\
160 & 3 & 1 & 0 & 0 & 0 & 0 & 0 & 1 & 0 & 0 & 0 & 0 & 0 & 0 & 0 \\
\midrule
161 & 3 & 0 & 6 & 0 & 0 & 0 & 0 & 0 & 0 & 0 & 0 & 0 & 0 & 0 & 0 \\
162 & 2 & 0 & 3 & 0 & 0 & 0 & 0 & 0 & 0 & 0 & 0 & 0 & 0 & 0 & 0 \\
163 & 2 & 0 & 2 & 0 & 0 & 0 & 0 & 0 & 0 & 0 & 0 & 0 & 0 & 0 & 0 \\
164 & 2 & 1 & 3 & 0 & 0 & 0 & 0 & 0 & 0 & 0 & 0 & 0 & 0 & 1 & 0 \\
165 & 2 & 0 & 0 & 0 & 0 & 0 & 0 & 0 & 0 & 0 & 0 & 0 & 0 & 0 & 0 \\
166 & 2 & 0 & 0 & 0 & 0 & 0 & 0 & 0 & 0 & 0 & 0 & 0 & 0 & 0 & 0 \\
167 & 2 & 0 & 0 & 0 & 1 & 0 & 0 & 0 & 0 & 0 & 0 & 0 & 0 & 0 & 0 \\
168 & 2 & 0 & 0 & 0 & 0 & 0 & 0 & 0 & 0 & 0 & 0 & 0 & 0 & 0 & 0 \\
169 & 2 & 0 & 0 & 0 & 0 & 0 & 0 & 0 & 0 & 0 & 0 & 0 & 0 & 0 & 0 \\
170 & 2 & 1 & 2 & 0 & 0 & 0 & 0 & 0 & 0 & 0 & 0 & 0 & 0 & 0 & 0 \\
\midrule
171 & 2 & 2 & 0 & 0 & 0 & 0 & 0 & 0 & 0 & 0 & 0 & 0 & 0 & 0 & 0 \\
172 & 2 & 0 & 2 & 0 & 0 & 0 & 0 & 0 & 0 & 0 & 0 & 0 & 0 & 0 & 0 \\
173 & 2 & 1 & 0 & 0 & 0 & 0 & 0 & 1 & 0 & 0 & 0 & 0 & 0 & 0 & 0 \\
174 & 2 & 0 & 1 & 0 & 0 & 0 & 0 & 0 & 0 & 0 & 0 & 0 & 0 & 0 & 0 \\
175 & 2 & 0 & 1 & 0 & 0 & 0 & 0 & 0 & 0 & 0 & 0 & 0 & 0 & 0 & 0 \\
176 & 2 & 1 & 0 & 0 & 0 & 0 & 0 & 0 & 0 & 0 & 0 & 0 & 0 & 0 & 0 \\
177 & 2 & 0 & 1 & 0 & 0 & 0 & 0 & 0 & 0 & 0 & 0 & 0 & 0 & 0 & 0 \\
178 & 1 & 0 & 0 & 0 & 0 & 0 & 0 & 0 & 0 & 0 & 0 & 0 & 0 & 0 & 0 \\
179 & 1 & 0 & 0 & 0 & 0 & 0 & 0 & 0 & 0 & 0 & 0 & 0 & 0 & 0 & 0 \\
180 & 1 & 0 & 0 & 0 & 0 & 0 & 0 & 0 & 0 & 0 & 0 & 0 & 0 & 0 & 0 \\
\midrule
181 & 1 & 0 & 0 & 0 & 0 & 0 & 0 & 0 & 0 & 0 & 0 & 0 & 0 & 0 & 0 \\
182 & 1 & 0 & 0 & 0 & 0 & 0 & 0 & 0 & 0 & 0 & 0 & 0 & 0 & 0 & 0 \\
183 & 1 & 0 & 0 & 0 & 0 & 0 & 0 & 0 & 0 & 0 & 0 & 0 & 0 & 0 & 0 \\
184 & 1 & 1 & 0 & 0 & 0 & 0 & 0 & 0 & 0 & 0 & 0 & 0 & 0 & 0 & 0 \\
185 & 1 & 0 & 2 & 0 & 0 & 0 & 0 & 0 & 0 & 0 & 0 & 0 & 0 & 0 & 0 \\
186 & 1 & 0 & 0 & 0 & 0 & 0 & 0 & 0 & 0 & 0 & 0 & 0 & 0 & 0 & 0 \\
187 & 1 & 0 & 0 & 0 & 0 & 0 & 0 & 0 & 0 & 0 & 0 & 0 & 0 & 0 & 0 \\
188 & 1 & 1 & 0 & 0 & 0 & 0 & 0 & 0 & 0 & 0 & 0 & 0 & 0 & 0 & 0 \\
189 & 1 & 0 & 0 & 0 & 0 & 0 & 0 & 1 & 0 & 0 & 0 & 0 & 0 & 0 & 0 \\
190 & 1 & 1 & 0 & 0 & 0 & 0 & 0 & 0 & 0 & 0 & 0 & 0 & 0 & 0 & 0 \\
\midrule
191 & 1 & 0 & 2 & 0 & 0 & 0 & 0 & 0 & 0 & 0 & 0 & 0 & 0 & 0 & 0 \\
192 & 1 & 0 & 0 & 0 & 0 & 0 & 0 & 0 & 0 & 0 & 0 & 0 & 0 & 0 & 0 \\
193 & 1 & 0 & 1 & 0 & 0 & 0 & 0 & 0 & 0 & 0 & 0 & 0 & 0 & 0 & 0 \\
194 & 1 & 0 & 0 & 0 & 0 & 0 & 0 & 0 & 0 & 0 & 0 & 0 & 0 & 0 & 0 \\
195 & 1 & 0 & 1 & 0 & 0 & 0 & 0 & 0 & 0 & 0 & 0 & 0 & 0 & 0 & 0 \\
196 & 1 & 0 & 2 & 0 & 0 & 0 & 0 & 0 & 0 & 0 & 0 & 0 & 0 & 0 & 0 \\
197 & 1 & 0 & 1 & 0 & 0 & 0 & 0 & 0 & 0 & 0 & 0 & 0 & 0 & 0 & 0 \\
198 & 1 & 0 & 1 & 0 & 0 & 0 & 0 & 0 & 0 & 0 & 0 & 0 & 0 & 0 & 0 \\
199 & 0 & 0 & 0 & 0 & 0 & 0 & 0 & 0 & 0 & 0 & 0 & 0 & 0 & 0 & 0 \\
200 & 0 & 0 & 0 & 0 & 0 & 0 & 0 & 0 & 0 & 0 & 0 & 0 & 0 & 0 & 0 \\
\midrule
201 & 0 & 0 & 0 & 0 & 0 & 0 & 0 & 0 & 0 & 0 & 0 & 0 & 0 & 0 & 0 \\
202 & 0 & 1 & 0 & 0 & 0 & 0 & 0 & 0 & 0 & 0 & 0 & 0 & 0 & 0 & 0 \\
203 & 0 & 0 & 1 & 0 & 0 & 0 & 0 & 0 & 0 & 0 & 0 & 0 & 0 & 0 & 0 \\
204 & 0 & 0 & 0 & 0 & 0 & 0 & 0 & 0 & 0 & 0 & 0 & 0 & 0 & 0 & 0 \\
205 & 0 & 0 & 0 & 0 & 0 & 0 & 0 & 0 & 0 & 0 & 0 & 0 & 0 & 0 & 0 \\
206 & 0 & 0 & 0 & 0 & 0 & 0 & 0 & 0 & 0 & 0 & 0 & 0 & 0 & 0 & 0 \\
207 & 0 & 0 & 0 & 0 & 0 & 0 & 0 & 0 & 0 & 0 & 0 & 0 & 0 & 0 & 0 \\
208 & 0 & 1 & 2 & 0 & 0 & 0 & 0 & 0 & 0 & 0 & 0 & 0 & 0 & 0 & 0 \\
209 & 0 & 0 & 1 & 0 & 0 & 0 & 0 & 0 & 0 & 0 & 0 & 0 & 0 & 0 & 0 \\
210 & 0 & 0 & 0 & 0 & 0 & 0 & 0 & 0 & 0 & 0 & 0 & 0 & 0 & 0 & 0 \\
\bottomrule
\end{longtable}

\end{document}